\documentclass[11pt]{article}
\usepackage[T1]{fontenc}
\usepackage{nicefrac}
\usepackage{microtype}
\usepackage{natbib}
\usepackage[dvipsnames]{xcolor}
\usepackage{fullpage}
\usepackage{tabularx}
\usepackage{enumitem}
\usepackage{algorithm}
\usepackage{algpseudocode}
\usepackage{mmll}
\usepackage[colorlinks,
            linkcolor=red,
            anchorcolor=blue,
            citecolor=blue,
            pagebackref=true
           ]{hyperref}
\usepackage[nameinlink]{cleveref}
\numberwithin{equation}{section}
\makeatletter
\providecommand{\theHALG@line}{\arabic{algorithm}.\arabic{ALG@line}}
\makeatother

\newcommand{\algname}{\text{SignRR}}
\newcommand{\algnameVR}{\text{SignRVR}}

\title{Convergence of Sign-based Random Reshuffling Algorithms for Nonconvex Optimization}

\author{
    Zhen Qin\thanks{The Ohio State University; e-mail: \texttt{qin.681@osu.edu}}~\footnotemark[3]~,
    \quad
    Zhishuai Liu\thanks{Duke University; e-mail: \texttt{zhishuai.liu@duke.edu}}~\thanks{Equal contribution}~,
    \quad
    Pan Xu\thanks{Duke University; e-mail: \texttt{pan.xu@duke.edu}}
}
\date{}
\allowdisplaybreaks
\begin{document}

\maketitle
\begin{abstract}
  signSGD is attractive in nonconvex optimization because it communicates sign-valued rather than full-precision gradients. Several standard analyses assume independent stochastic-gradient samples, whereas a common finite-sum implementation reshuffles the data and processes them sequentially. We study this variant, signSGD with random reshuffling (SignRR), and show that reshuffling does not in general repair the bias created by discarding gradient magnitudes. In particular, on a one-dimensional two-component strongly convex quadratic, the expected gradient norm at every SignRR inner iterate equals $1/2$. We complement this impossibility result with an alignment-explicit finite-time bound $O(\log(nT)/\sqrt{nT}+\varepsilon_{\mathrm{align}})$, where $\varepsilon_{\mathrm{align}}$ measures the averaged loss of descent caused by component-sign misalignment. A horizon-tuned constant stepsize improves the vanishing term to $O(1/\sqrt{nT})$, and a remaining-set alignment condition yields a residual-free $O(1/\sqrt{nT})$ guarantee. The alignment term is upper bounded by twice the averaged mean absolute gradient error and, in turn, by twice an averaged coordinatewise conditional root-mean-square error. As a variance-reduced alternative, we analyze SignRVR, which signs an SVRG estimator anchored at the beginning of every epoch. A pathwise argument gives a residual-free guarantee with an $O(\sqrt{d/T})$ averaged $\ell_1$-stationarity bound.

\end{abstract}

\section{Introduction}
We study the following optimization problem 
\begin{align}
    \min_{\xb\in\RR^d}\quad f(\xb),
    \qquad
    f(\xb):=\frac{1}{n}\sum_{i=0}^{n-1}f_i(\xb).
    \label{eq:problem}
\end{align}
where each component $f_i$ is smooth and possibly nonconvex and may represent the loss on one training example. Global minimization can be computationally intractable even for structured nonconvex objectives; for example, many tensor-optimization problems are NP-hard \citep{hillar2013most}. Nonconvex first-order theory therefore commonly targets an $\varepsilon$-approximate stationary point $\xb\in\RR^d$ satisfying $\lVert\nabla f(\xb)\rVert_2\leq\varepsilon$ for a prescribed $\varepsilon>0$. Full-gradient descent is a basic method for this objective, but each update evaluates all $n$ component gradients. This cost can be prohibitive for a large dataset.

To address the cost of large datasets, many stochastic first-order and variance-reduced methods use one or a mini-batch of component gradients per update \citep{robbins1951stochastic,bottou2009curiously,bottou2012stochastic,reddi2016stochastic,allen-zhu2016variance,lei2017nonconvex,nguyen2017sarah,nguyen2017stochastic,fang2018spider,zhou2020stochastic}. Their sampling schemes and gradient estimators vary, but their per-update component cost can be much smaller than that of a full gradient.

For distributed training of large models, communicating full-precision gradients can also be expensive. Quantization and sparsification \citep{seide20141bit,zhang2017zipml,lin2017deep} are common forms of gradient compression \citep{alistarh2017qsgd,wen2017terngrad,wang2018atomo,khirirat2018distributed}. Sign-based update rules have been widely studied because of their simplicity \citep{carlson2015stochastic,liu2019signsgd,balles2018dissecting,bernstein2018signsgd}. In distributed variants, workers can transmit gradient signs rather than full-precision coordinates \citep{seide20141bit,strom2015scalable,bernstein2019signsgd}. This reduces gradient-communication bandwidth, although it does not reduce the memory needed to store the model itself.
Recent analyses characterize when sign-based SGD converges despite the bias introduced by the sign map \citep{bernstein2018signsgd,bernstein2019signsgd,safaryan2021stochastic}.

Several analyses of sign-based stochastic gradient descent (signSGD) use independent stochastic-gradient oracles or impose explicit sign-success conditions \citep{bernstein2018signsgd,bernstein2019signsgd,safaryan2021stochastic,chzhen2023signsvrg}. This differs from the finite-sum implementation that independently permutes the dataset at the beginning of each epoch and processes the resulting order sequentially. Random reshuffling (RR) is easy to implement and often performs well empirically \citep{bottou2012stochastic,recht2012beneath,recht2013parallel,sun2019optimization,gurbuzbalaban2021why}, but its theoretical advantage over with-replacement SGD depends on the objective class, stepsize, performance criterion, and epoch budget \citep{mishchenko2020random,nguyen2021unified,safran2020how,safran2021random}. The interaction between RR and the nonlinear sign map motivates our central question:
\begin{center}
 \emph{Does signSGD with without-replacement sampling admit a finite-time $\varepsilon$-stationarity guarantee?}
\end{center}
Our answer for vanilla SignRR is negative without further structure. The primary obstruction is the biased direction created by applying the sign map before averaging; without-replacement dependence is a secondary analytical issue. We isolate this distinction and then study conditions and an algorithmic mechanism that control the sign error.

\textbf{Our main contributions} are summarized as follows.
 \begin{itemize}[leftmargin=*]
    \item We study signSGD with random reshuffling, namely $\algname$,
    which updates the parameter using the sign of the stochastic gradient and reads the dataset sequentially. We use an averaged sign-alignment deficit $\varepsilon_{\mathrm{align}}$ to capture the exact loss of first-order descent caused by biased component signs. We prove that this term cannot be omitted from a uniform guarantee covering the stated diminishing schedule and problem class by giving a smooth strongly convex problem on which $\EE\lVert\nabla f\rVert_1=1/2$ at every SignRR inner iterate. Without an alignment assumption, our bounds are $O(\log(nT)/\sqrt{nT}+\varepsilon_{\mathrm{align}})$ for the stated diminishing schedule and $O(1/\sqrt{nT}+\varepsilon_{\mathrm{align}})$ for a horizon-tuned constant step. Under a conditional remaining-set alignment condition, the latter becomes a residual-free $O(1/\sqrt{nT})$ rate. We also give weaker bounds in terms of averaged mean absolute and coordinatewise root-mean-square gradient errors.
    This provides a finite-time analysis of sign-based SGD under repeated random reshuffling for nonconvex finite-sum optimization.
    \item Inspired by SVRG and SignSVRG \citep{johnson2013accelerating,chzhen2023signsvrg}, we analyze a distinct RR variant, $\algnameVR$, that directly signs an epoch-anchored SVRG control variate without dithering. For arbitrary positive deterministic stepsizes, we prove a pathwise bound controlled by the squared total stepsize within each epoch. For a horizon-tuned constant step, this yields an averaged $\ell_1$-stationarity guarantee of order $O(\sqrt{d/T})$, where $T$ is the number of epochs. The proof is valid for every component order and requires neither bounded variance nor a coordinatewise sign-success assumption.
\end{itemize}   

\paragraph{Notation} Let $n,d,T\in\NN^+$. For $k\in\NN^+$, let $[k]:=\{0,1,\ldots,k-1\}$. We write $[\xb]_j$ for the $j$-th coordinate of $\xb$, $\lVert\xb\rVert_1=\sum_{j=1}^d|[\xb]_j|$, and $\lVert\xb\rVert_2=\sqrt{\xb^\top\xb}$. Unless ranges are shown explicitly, $\sum_{t,i}$ and $\min_{t,i}$ mean $\sum_{t=0}^{T-1}\sum_{i=0}^{n-1}$ and $\min_{0\leq t<T,\,0\leq i<n}$, respectively. For functions $g$ and $h$, the notation $g(T)=O(h(T))$ hides constants independent of $T$. For an event $A$, $\ind\{A\}$ denotes its indicator.

\section{Related Work}
We summarize two lines of work that are most relevant.

\paragraph{Sign-based SGD}
Sign-based algorithms are attractive for their simple bounded updates \citep{riedmiller1993direct,balles2018dissecting,bernstein2018signsgd}, while distributed variants enable sign-only communication \citep{seide20141bit,strom2015scalable,bernstein2019signsgd}. For nonzero coordinates the sign is binary; exact zeros require an agreed encoding convention. The nonconvex analysis of \cite{bernstein2018signsgd} assumes an independent unbiased stochastic-gradient oracle with coordinatewise variance bounds. Its main theorem uses $K$ updates with mini-batch size $K$, hence $O(K^2)$ stochastic-gradient calls, to control coordinatewise wrong-sign probabilities. The resulting geometry depends on the $\ell_1$ norm of the gradient as well as coordinatewise curvature and noise. The same work studies Signum and majority-vote aggregation, but its guarantees rely on the stated noise and batch assumptions.

Two subsequent directions make the sign bias explicit. \cite{safaryan2021stochastic} uses coordinatewise success-probability bounds and allows biased estimators; such bounds remove the particular large-batch construction but impose a direct sign-correctness condition. \cite{karimireddy19a} gives convex counterexamples for plain signSGD and shows that error feedback can repair biased compression by retaining discarded information. Momentum-based gradient estimators provide another route to small-batch guarantees for normalized or generalized sign methods \citep{cutkosky2020momentum,crawshaw2022robustness}.

For finite-sum optimization, \cite{chzhen2023signsvrg} proposes SignSGD+ and SignSVRG. SignSGD+ uses calibrated uniform dither so that, under its bounded-gradient assumption, the expected signed direction is proportional to the gradient; SignSVRG combines this device with an SVRG control variate. More recent variance-reduced sign methods improve stochastic and finite-sum rates under their respective assumptions \citep{jiang2024efficient}. These works illustrate that residual-free guarantees can be obtained by controlling or correcting the signed estimator. In distributed problems, majority vote introduces an additional heterogeneity issue because a vote over worker signs need not align with the magnitude-weighted global gradient \citep{safaryan2021stochastic,jin2023magnitude}.

\paragraph{Random reshuffling}
RR uses every component exactly once in a uniformly random order in each epoch. It often improves empirical training behavior, but theoretical comparisons with with-replacement SGD depend on the objective class, stepsize, condition number, epoch budget, and output criterion. The strongly convex literature gives refined upper and lower bounds for repeated reshuffling, shuffle-once, and incremental orders \citep{gurbuzbalaban2021why,nagaraj2019sgd,haochen2019random,safran2020how,rajput2020closing,ahn2020sgd}. In particular, lower-bound results show that the advantage of reshuffling is regime-dependent and may require sufficiently many epochs relative to conditioning \citep{safran2020how,safran2021random}.
For componentwise convex and smooth finite sums, \cite{liu2026random} proves an expected-suboptimality rate bound for a weighted-average RR output that dominates the corresponding SGD rate bound for stepsizes up to order $1/L_{\max}$, where $L_{\max}$ is its componentwise smoothness constant; it does not cover nonconvex objectives or component-sign updates.

For smooth nonconvex finite sums, \cite{mishchenko2020random} develops an analysis based on shuffling variance, while \cite{nguyen2021unified} gives a unified treatment of arbitrary shuffling and uniformly randomized reshuffling. The latter reports stationarity rates of order $T^{-2/3}$ for general shuffling and $n^{-1/3}T^{-2/3}$ in a randomized setting under its assumptions, where $T$ counts epochs. These criteria are based on squared $\ell_2$ gradient norms and should not be compared directly with an expected unsquared $\ell_1$ gradient norm without an explicit conversion. Variance reduction can also be combined with RR and yields faster guarantees in structured regimes \citep{haochen2021variance,malinovsky2023random}.

The key analytical device in ordinary RR is that full-precision component gradients cancel correctly at a fixed epoch anchor. SignRR does not inherit this identity because it averages component signs rather than component gradients. Moreover, a theorem for random reshuffling does not automatically cover shuffle-once or cyclic orders; these are distinct sampling schemes requiring their own argument. Our focus on RR addresses the common repeated-permutation implementation, while \Cref{prop:signrr-impossibility} shows that even this randomized ordering does not in general repair magnitude loss by itself.

\section{What Can and Cannot Be Guaranteed for SignRR}
Sign-based updates are attractive because they communicate sign-valued coordinates and can be viewed as steepest-descent steps in $\ell_\infty$ geometry \citep{bernstein2018signsgd,balles2018dissecting}. Random reshuffling (RR), meanwhile, often improves ordinary finite-sum gradient methods by exploiting the cancellation identity $\sum_{r=0}^{n-1}\nabla f_r(\xb)=n\nabla f(\xb)$. SignRR combines the two ideas, but the nonlinear sign map destroys this identity:
\begin{align*}
    \frac1n\sum_{r=0}^{n-1}\sign(\nabla f_r(\xb))
    \quad\text{need not align with}\quad
    \sign(\nabla f(\xb)).
\end{align*}
Consequently, processing every component once per epoch does not by itself remove stochastic sign bias. This section gives an alignment-explicit descent inequality, a strongly convex counterexample showing that lower boundedness and smoothness alone do not yield a uniform residual-free guarantee covering the stated diminishing schedule, and a sufficient alignment condition under which SignRR does converge.

\subsection{The SignRR Algorithm}
The SignRR algorithm is formally defined in \Cref{alg:signrr}. At each epoch $t$, it draws a permutation $\pi^t=(\pi_0^t,\ldots,\pi_{n-1}^t)$ uniformly from all permutations of $[n]$, independently of the permutations used in earlier epochs. It then updates $\xb$ sequentially in this order. The sign map is applied coordinatewise, and each component index is used exactly once within an epoch.

\begin{algorithm}
\caption{Sign-based Random Reshuffling ($\algname$)\label{alg:signrr}}
\begin{algorithmic}
\Require Positive stepsizes $\{\gamma_t^i:0\leq t<T,\ 0\leq i<n\}$, initialization $\xb_0\in\RR^d$, number of epochs $T$.
\For{$t=0,\ldots,T-1$}
\State Independently draw a uniform random permutation $\pi^t=(\pi_0^t,\ldots,\pi_{n-1}^t)$ of $[n]$ and set $\xb_t^0=\xb_t$.
\For{$i=0,\ldots,n-1$}
\State $\xb^{i+1}_t = \xb^i_t- \gamma_t^i \sign(\nabla f_{\pi^t_i}(\xb_t^i))$.
\EndFor
\State $\xb_{t+1} = \xb^n_t$.
\EndFor
\end{algorithmic}
\end{algorithm}

\subsection{An alignment-explicit bound under smoothness}
To analyze SignRR, we consider the nonconvex finite-sum problem defined in \eqref{eq:problem} and adopt standard assumptions from prior work on sign-based methods \citep{bernstein2018signsgd,bernstein2019signsgd,safaryan2021stochastic,chzhen2023signsvrg}.
\begin{assumption}\label{ass:lower-bound}
There exists $f_*\in\RR$ such that $f(\xb)\geq f_*$ for every $\xb\in\RR^d$.
\end{assumption}
We also assume component smoothness.
\begin{assumption}\label{ass:smoothness}
For some $L>0$, every component $f_i$ has an $L$-Lipschitz gradient: $\lVert\nabla f_i(\xb)-\nabla f_i(\yb)\rVert_2\leq L\lVert\xb-\yb\rVert_2$ for all $\xb,\yb\in\RR^d$. Consequently, their average $f$ is also $L$-smooth and satisfies $f(\xb)\leq f(\yb)+\langle\nabla f(\yb),\xb-\yb\rangle+\frac{L}{2}\lVert\xb-\yb\rVert_2^2$.
\end{assumption}
The scalar smoothness assumption is sufficient for every argument below. A coordinatewise smoothness model can retain more detailed anisotropic constants, but we do not use that structure here.

We use the mathematical convention $\sign(0)=0$; thus the update alphabet is ternary when exact zero coordinates occur. For $\gb,\vb\in\RR^d$, define the sign-alignment deficit
\begin{align}
    \mathsf A(\gb,\vb)
    :=\lVert\gb\rVert_1-\langle \gb,\sign(\vb)\rangle
    =\sum_{j=1}^d |g_j|\bigl(1-\sign(g_j)\sign(v_j)\bigr).
    \label{eq:alignment-deficit}
\end{align}
This nonnegative quantity is exactly the amount of first-order descent lost by using $\sign(\vb)$ instead of $\sign(\gb)$. For integers $m\geq1$ and $r>0$, let $H_{m,r}:=\sum_{k=1}^m k^{-r}$, abbreviate $H_m:=H_{m,1}$, set $H_0:=0$, and let $N:=nT$. Define
\begin{align}
    \varepsilon_{\mathrm{align}}
    :=\frac{\sum_{t=0}^{T-1}\sum_{i=0}^{n-1}\gamma_t^i
    \EE\!\left[\mathsf A\!\left(\nabla f(\xb_t^i),\nabla f_{\pi_i^t}(\xb_t^i)\right)\right]}
    {\sum_{t=0}^{T-1}\sum_{i=0}^{n-1}\gamma_t^i}.
    \label{eq:average-alignment-deficit}
\end{align}
\begin{theorem}\label{thm:signrr}
Under \Cref{ass:lower-bound,ass:smoothness}, set
$\gamma_t^i=\gamma_0/\sqrt{nt+i+1}$, where $\gamma_0>0$. The iterates of \Cref{alg:signrr} satisfy
\begin{align}
    \min_{\substack{0\leq i\leq n-1\\0\leq t\leq T-1}}
    \EE\!\left[\lVert\nabla f(\xb_t^i)\rVert_1\right]
    &\leq
    \frac{\sum_{t=0}^{T-1}\sum_{i=0}^{n-1}\gamma_t^i
    \EE\!\left[\lVert\nabla f(\xb_t^i)\rVert_1\right]}
    {\sum_{t=0}^{T-1}\sum_{i=0}^{n-1}\gamma_t^i} \notag\\
    &\leq
    \frac{f(\xb_0)-f_*}{\gamma_0 H_{N,1/2}}
    +\frac{dL\gamma_0}{2}\frac{H_N}{H_{N,1/2}}
    +\varepsilon_{\mathrm{align}}.
    \label{eq:signrr-alignment-bound}
\end{align}
Consequently,
\begin{align}
    \min_{t,i}\EE\!\left[\lVert\nabla f(\xb_t^i)\rVert_1\right]
    \leq
    \frac{f(\xb_0)-f_*}{2\gamma_0(\sqrt{N+1}-1)}
    +\frac{dL\gamma_0(1+\log N)}{4(\sqrt{N+1}-1)}
    +\varepsilon_{\mathrm{align}}.
    \label{eq:signrr-explicit-bound}
\end{align}
\end{theorem}
An implementable output matching the weighted-average bound is obtained by
sampling $(t,i)$ independently of the algorithmic randomness, with probability
$\gamma_t^i/\sum_{s=0}^{T-1}\sum_{r=0}^{n-1}\gamma_s^r$ and returning $\xb_t^i$; its expected
gradient norm is the weighted average displayed in
\eqref{eq:signrr-alignment-bound}.
To compare the alignment term with an RMS surrogate, define the averaged mean absolute error
\begin{align*}
    \delta_{\mathrm{abs}}
    :=\frac{\sum_{t,i}\gamma_t^i
    \EE\lVert\nabla f_{\pi_i^t}(\xb_t^i)-\nabla f(\xb_t^i)\rVert_1}
    {\sum_{t,i}\gamma_t^i},
\end{align*}
and define the nonnegative conditional RMS coordinate
\begin{align*}
    [\mathbf e_t^i]_j
    :=\sqrt{\EE\!\left[
    \bigl([\nabla f_{\pi_i^t}(\xb_t^i)]_j
    -[\nabla f(\xb_t^i)]_j\bigr)^2
    \,\middle|\,\xb_t^i\right]}
\end{align*}
and
\begin{align*}
    \sigma_{\mathrm{rms}}:=
    \frac{\sum_{t,i}\gamma_t^i\EE\lVert\mathbf e_t^i\rVert_1}
    {\sum_{t,i}\gamma_t^i}.
\end{align*}
Then the absolute-error bound \eqref{eq:alignment-absolute-bound} in
\Cref{lem:alignment-deficit} and conditional Cauchy--Schwarz give
\begin{align}
    \varepsilon_{\mathrm{align}}\leq 2\delta_{\mathrm{abs}}\leq 2\sigma_{\mathrm{rms}}.
    \label{eq:alignment-error-hierarchy}
\end{align}
Thus \eqref{eq:signrr-alignment-bound} also yields an RMS-error bound, while the alignment term avoids this surrogate's magnitude-error slack whenever the stochastic gradient has a large magnitude error without a corresponding sign error. The sign-descent geometry here naturally yields an $\ell_1$-stationarity criterion; because $\lVert\nabla f\rVert_2\leq\lVert\nabla f\rVert_1$, the same bound also controls $\ell_2$-stationarity.

The inequality \eqref{eq:one-step-alignment-descent} in
\Cref{lem:one-step-signed-descent} is pathwise
and does not use independence or even the fact that the indices form a
permutation; the displayed bounds follow after taking expectations. The
result therefore isolates the precise first-order alignment cost, but does
not yet exploit RR. The RMS quantity $\sigma_{\mathrm{rms}}$ can be loose:
magnitude error is irrelevant whenever it does not change a sign.

\begin{corollary}[Constant and optimized stepsizes]\label{cor:signrr-constant}
Using a constant stepsize $\gamma_t^i=\gamma>0$ and defining $\varepsilon_{\mathrm{align}}^{\mathrm{const}}$ by \eqref{eq:average-alignment-deficit} with constant weights, the same argument yields
\begin{align*}
    \frac{1}{nT}\sum_{t=0}^{T-1}\sum_{i=0}^{n-1}
    \EE\lVert\nabla f(\xb_t^i)\rVert_1
    \leq \frac{f(\xb_0)-f_*}{nT\gamma}+\frac{dL\gamma}{2}
    +\varepsilon_{\mathrm{align}}^{\mathrm{const}}.
\end{align*}
Writing $\Delta_0:=f(\xb_0)-f_*$, if $\Delta_0>0$ and
\begin{align*}
    \gamma=\sqrt{\frac{2\Delta_0}{dLN}},
\end{align*}
then the two vanishing terms are optimized and
\begin{align}
    \frac{1}{N}\sum_{t,i}\EE\lVert\nabla f(\xb_t^i)\rVert_1
    \leq \sqrt{\frac{2dL\Delta_0}{N}}
    +\varepsilon_{\mathrm{align}}^{\mathrm{const}}.
    \label{eq:signrr-optimized-constant}
\end{align}
Thus a horizon-tuned constant step removes the logarithm from the vanishing part of \eqref{eq:signrr-explicit-bound}. The alignment term can again be upper bounded by twice the corresponding mean absolute or coordinatewise RMS error.
\end{corollary}

\subsection{Why an alignment term is unavoidable}
The next result shows why the horizon-dependent term $\varepsilon_{\mathrm{align}}$ cannot simply be omitted from a general SignRR bound. Even on a one-dimensional strongly convex quadratic, RR can cancel the component \emph{signs} while the component gradient magnitudes---and hence the full gradient---do not cancel.

\begin{proposition}[Smooth strongly convex nonconvergence example]
\label{prop:signrr-impossibility}
Let $d=1$, $n=2$, and
\begin{align*}
    f_0(x)=\frac12(x+2)^2,
    \qquad
    f_1(x)=\frac12(x-1)^2,
    \qquad
    f(x)=\frac12\bigl(f_0(x)+f_1(x)\bigr).
\end{align*}
Every component and the average objective are $1$-smooth and $1$-strongly convex. Initialize $x_0=0$ and use
$\gamma_t^i=\gamma_0/\sqrt{2t+i+1}$ with $0<\gamma_0\leq1/4$. Then, for every epoch $t\geq0$ and both $i\in\{0,1\}$,
\begin{align}
    \EE\bigl[|f'(x_t^i)|\bigr]=\frac12.
    \label{eq:signrr-counterexample-gradient}
\end{align}
In particular, the expected gradient norm does not approach zero as the number of epochs grows.
\end{proposition}

This counterexample rules out a uniform residual-free SignRR guarantee that covers the stated schedule and problem class under only lower boundedness and smoothness, even if component convexity and strong convexity are added. The obstruction is the biased sign field, not the statistical dependence of RR. This is consistent with known convex counterexamples for signSGD and the role of error feedback in repairing biased compression \citep{karimireddy19a}.

A direct aggregate sufficient condition is a relative control of the
alignment deficit.
\begin{proposition}[Aggregate relative alignment]\label{prop:aggregate-alignment}
Under \Cref{ass:lower-bound,ass:smoothness}, suppose that for some $\rho\in(0,1]$ and
$B_N\geq0$,
\begin{align}
    \sum_{t,i}\gamma_t^i\EE\mathsf A(\gb_t^i,\vb_t^i)
    \leq (1-\rho)\sum_{t,i}\gamma_t^i
      \EE\lVert\gb_t^i\rVert_1+B_N,
    \label{eq:aggregate-relative-alignment}
\end{align}
where $\gb_t^i=\nabla f(\xb_t^i)$ and
$\vb_t^i=\nabla f_{\pi_i^t}(\xb_t^i)$. Then
\begin{align}
    \min_{t,i}\EE\lVert\nabla f(\xb_t^i)\rVert_1
    \leq
    \frac{f(\xb_0)-f_*+\frac{dL}{2}\sum_{t,i}(\gamma_t^i)^2+B_N}
         {\rho\sum_{t,i}\gamma_t^i}.
    \label{eq:aggregate-alignment-bound}
\end{align}
\end{proposition}
Condition \eqref{eq:aggregate-relative-alignment} is aggregate but
trajectory- and horizon-dependent. The following condition is stronger, more
interpretable, and explicitly adapted to RR.

\subsection{A residual-free guarantee under remaining-set alignment}
Although \Cref{alg:signrr} draws $\pi^t$ at the beginning of epoch $t$,
we expose its entries sequentially for the analysis; this deferred-decision
view has the same distribution as drawing the full permutation at once. Let
$\mathcal H_t:=\sigma(\pi^0,\ldots,\pi^{t-1})$ denote the history through
the end of epoch $t-1$, with $\mathcal H_0$ the trivial sigma-field, and
define
\begin{align*}
    \mathcal F_t^i
    &:=\mathcal H_t\vee
      \sigma(\pi_r^t:0\leq r<i),\\
    \mathcal R_t^i
    &:=[n]\setminus\{\pi_r^t:0\leq r<i\},
\end{align*}
with an empty prefix when $i=0$. Thus $\mathcal F_t^i$ is the information
exposed in this analysis immediately before the $i$-th inner update:
exactly $i$ component indices have been exposed and used, while $n-i$
indices remain. Both
$\xb_t^i$ and $\mathcal R_t^i$ are $\mathcal F_t^i$-measurable, and,
conditionally on $\mathcal F_t^i$, $\pi_i^t$ is uniform on
$\mathcal R_t^i$. Consequently, the conditional mean signed direction of
the next update is
\begin{align}
    \mathbf{s}_t^i
    :=\EE\!\left[\sign(\nabla f_{\pi_i^t}(\xb_t^i))\mid\mathcal F_t^i\right]
    =\frac{1}{n-i}\sum_{r\in\mathcal R_t^i}
      \sign(\nabla f_r(\xb_t^i)).
    \label{eq:remaining-sign-field}
\end{align}
The average over $\mathcal R_t^i$ appears precisely because, given the
exposed prefix, each unused index is equally likely to be $\pi_i^t$.

\begin{assumption}[Remaining-set sign alignment]\label{ass:remaining-alignment}
There is a constant $\rho\in(0,1]$ such that, almost surely for every
$t\in[T]$ and $i\in[n]$,
\begin{align}
    \left\langle\nabla f(\xb_t^i),\mathbf{s}_t^i\right\rangle
    \geq \rho\lVert\nabla f(\xb_t^i)\rVert_1.
    \label{eq:remaining-alignment}
\end{align}
\end{assumption}

This condition asks for the conditional \emph{average} direction among unused components to be a descent direction; it does not require all component signs to agree at every common point. It is nevertheless strong: when $i=n-1$, the remaining set is a singleton, so the last unused component must itself satisfy the alignment inequality at the realized iterate. A conditional coordinatewise success-probability bound implies the condition when the signed correctness margin is at least $\rho$, but \eqref{eq:remaining-alignment} also permits errors on coordinates carrying little full-gradient mass.
By the tower property, \Cref{ass:remaining-alignment} implies
\eqref{eq:aggregate-relative-alignment} with $B_N=0$.

\begin{theorem}[Residual-free SignRR convergence]\label{thm:signrr-aligned}
Under \Cref{ass:lower-bound,ass:smoothness,ass:remaining-alignment}, arbitrary positive deterministic stepsizes satisfy
\begin{align}
    \min_{t,i}\EE\lVert\nabla f(\xb_t^i)\rVert_1
    \leq
    \frac{f(\xb_0)-f_*+\frac{dL}{2}\sum_{t,i}(\gamma_t^i)^2}
         {\rho\sum_{t,i}\gamma_t^i}.
    \label{eq:signrr-aligned-general}
\end{align}
For the diminishing stepsizes of \Cref{thm:signrr}, the right-hand side is
\begin{align}
    \frac{f(\xb_0)-f_*}{\rho\gamma_0H_{N,1/2}}
    +\frac{dL\gamma_0}{2\rho}\frac{H_N}{H_{N,1/2}}.
    \label{eq:signrr-aligned-diminishing}
\end{align}
For the optimized constant stepsize in \Cref{cor:signrr-constant}, it is
\begin{align}
    \sqrt{\frac{2dL(f(\xb_0)-f_*)}{\rho^2N}}.
    \label{eq:signrr-aligned-optimized}
\end{align}
\end{theorem}

\paragraph{Relation to prior analyses.}
The original signSGD analysis controls sign errors using independent unbiased mini-batch gradients with coordinatewise variance bounds and uses batch size proportional to the number of iterations \citep{bernstein2018signsgd}. Success-probability analyses replace that large-batch construction by an explicit sign-correctness assumption \citep{safaryan2021stochastic}; under RR, the corresponding assumption must be conditional on the unused set, as in \eqref{eq:remaining-alignment}, rather than copied from an i.i.d. oracle. Variance-reduced sign methods instead make the signed estimator track the full gradient \citep{chzhen2023signsvrg}. Error feedback takes a different route by retaining the compression error and can recover SGD-type guarantees for biased compressors \citep{karimireddy19a}.

Ordinary RR theory cannot be transferred through the sign map: its epoch-level cancellation uses gradient magnitudes, whereas SignRR averages component votes. Moreover, RR is not uniformly faster than with-replacement SGD across all assumptions and iteration budgets; established improvements are regime- and criterion-dependent \citep{safran2020how,safran2021random,mishchenko2020random,nguyen2021unified}. Therefore, under the assumptions of \Cref{thm:signrr}, the general conclusion available here is the alignment-explicit finite-time bound together with the impossibility result. Residual-free rates can instead be sought under additional conditions such as \Cref{ass:remaining-alignment} or with corrections such as variance reduction, momentum-based estimation, magnitude-aware updates, or error feedback, each under assumptions appropriate to that mechanism.

\section{Variance Reduction of SignRR}
\Cref{thm:signrr,prop:signrr-impossibility} shows that reshuffling alone does not in general eliminate the sign-alignment deficit. Inspired by SVRG and SignSVRG \citep{johnson2013accelerating,chzhen2023signsvrg}, we therefore analyze a distinct variance-reduced RR method, \algnameVR, that directly signs an SVRG control variate anchored at the beginning of each epoch, without the dither used by SignSVRG. The estimator is
\begin{align}
    \vb_t^i:=\nabla f_{\pi_i^t}(\xb_t^i)-\nabla f_{\pi_i^t}(\yb_t)+\nabla f(\yb_t),
    \qquad \yb_t:=\xb_t.
    \label{eq:signrvr-estimator}
\end{align}
The algorithm always takes the signed step. Its bounded motion inside an epoch is sufficient for the analysis.

\begin{algorithm}
\caption{Sign-based random reshuffling with variance reduction ($\algnameVR$)}
\label{alg:signrvr}
\begin{algorithmic}
\Require Positive stepsizes $\{\gamma_t^i:0\leq t<T,\ 0\leq i<n\}$, initialization $\xb_0\in\RR^d$, number of epochs $T$.
\For{$t=0,\ldots,T-1$}
\State Independently draw a uniform random permutation $\pi^t=(\pi_0^t,\ldots,\pi_{n-1}^t)$ of $[n]$.
\State Set $\xb_t^0=\xb_t$, $\yb_t=\xb_t$, and compute $\nabla f(\yb_t)$.
\For{$i=0,\ldots,n-1$}
\State Form $\vb_t^i$ according to \eqref{eq:signrvr-estimator}.
\State $\xb_t^{i+1}=\xb_t^i-\gamma_t^i\sign(\vb_t^i)$.
\EndFor
\State $\xb_{t+1}=\xb_t^n$.
\EndFor
\end{algorithmic}
\end{algorithm}

For each epoch, define its total and squared stepsizes by
\begin{align}
    S_t:=\sum_{i=0}^{n-1}\gamma_t^i,
    \qquad
    Q_t:=\sum_{i=0}^{n-1}(\gamma_t^i)^2,
    \qquad
    \Delta_0:=f(\xb_0)-f_*.
    \label{eq:signrvr-step-totals}
\end{align}

\begin{theorem}\label{thm:signrvr}
Under \Cref{ass:lower-bound,ass:smoothness}, \Cref{alg:signrvr} with arbitrary positive deterministic stepsizes satisfies
\begin{align}
\sum_{t=0}^{T-1}\sum_{i=0}^{n-1}\gamma_t^i
  \EE\lVert\nabla f(\xb_t^i)\rVert_1 &\leq
\Delta_0+2dL\sum_{t=0}^{T-1}S_t^2
-\frac{3dL}{2}\sum_{t=0}^{T-1}Q_t
\leq \Delta_0+2dL\sum_{t=0}^{T-1}S_t^2.
\label{eq:signrvr-general}
\end{align}
Consequently,
\begin{align}
\min_{t,i}\EE\lVert\nabla f(\xb_t^i)\rVert_1
\leq
\frac{\Delta_0+2dL\sum_{t=0}^{T-1}S_t^2-\frac{3dL}{2}\sum_{t=0}^{T-1}Q_t}
{\sum_{t=0}^{T-1}S_t}.
\label{eq:signrvr-min-general}
\end{align}
The first inequality in \eqref{eq:signrvr-general} follows by taking expectations of the corresponding pathwise inequality obtained by removing $\EE$ from its left-hand side; that underlying inequality holds for every sequence of component indices. In particular, let $\gamma_t^i=\gamma_0/\sqrt{nt+i+1}$. Then
\begin{align}
\min_{t,i}\EE\lVert\nabla f(\xb_t^i)\rVert_1
\leq
\frac{\Delta_0+2dLn\gamma_0^2\bigl(4+H_{T-1}\bigr)}
{\gamma_0H_{N,1/2}}.
\label{eq:signrvr-diminishing}
\end{align}
\end{theorem}

\begin{corollary}\label{cor:signrvr-constant}
Under the conditions of \Cref{thm:signrvr}, suppose $\gamma_t^i=\gamma>0$ for every $t,i$. Then
\begin{align}
\frac1{nT}\sum_{t=0}^{T-1}\sum_{i=0}^{n-1}
\EE\lVert\nabla f(\xb_t^i)\rVert_1
\leq
\frac{\Delta_0}{nT\gamma}+dL\left(2n-\frac32\right)\gamma.
\label{eq:signrvr-constant}
\end{align}
If $\Delta_0>0$, the choice
\begin{align}
\gamma=\sqrt{\frac{\Delta_0}{nT\,dL(2n-3/2)}}
\end{align}
gives
\begin{align}
\frac1{nT}\sum_{t=0}^{T-1}\sum_{i=0}^{n-1}
\EE\lVert\nabla f(\xb_t^i)\rVert_1
\leq
2\sqrt{\frac{dL\Delta_0}{T}\left(2-\frac{3}{2n}\right)}.
\label{eq:signrvr-optimized}
\end{align}
\end{corollary}

\begin{remark}
The bound is residual-free and uses no bounded-variance or sign-success assumption. The direct $\ell_1$ argument yields a coefficient of order $nd$ and, after optimizing a constant stepsize, the dimension factor $\sqrt{d/T}$. A looser coordinatewise-RMS derivation can introduce an extra factor $\sqrt d$, but that loss is not inherent to every RMS analysis. The result is order-agnostic, however: random reshuffling is allowed but is not what produces the rate, so this theorem does not establish a theoretical advantage of RR over other component orders.
\end{remark}

\begin{remark}
Each epoch requires a full gradient at the anchor and, unless anchor component gradients are cached, two component-gradient evaluations per inner step. Comparisons with ordinary RR or with-replacement SignSVRG must therefore account for oracle cost as well as the different stationarity norms and squared versus unsquared criteria. We make no claim that \eqref{eq:signrvr-optimized} is an RR acceleration result.
\end{remark}

\section{Conclusions}
We characterized the central obstruction in signSGD with random reshuffling: applying the sign map before averaging destroys the cancellation used by ordinary RR. Our alignment-explicit descent inequality isolates this obstruction as a sign-alignment deficit, and a smooth strongly convex counterexample proves that this term cannot be omitted from a uniform vanishing bound covering the stated diminishing schedule under smoothness and lower boundedness alone. A conditional remaining-set alignment assumption recovers a residual-free SignRR guarantee.

We also analyzed \algnameVR, which signs an SVRG estimator anchored once per epoch. A direct pathwise argument gives a residual-free bound for arbitrary component orders and a horizon-tuned $O(\sqrt{d/T})$ averaged $\ell_1$-stationarity guarantee. Because this proof is order-agnostic, it establishes a guarantee for the variance-reduced method but not an advantage specific to random reshuffling. The present theory intentionally makes no claim for momentum or distributed majority-vote variants; correct guarantees for those methods require additional estimator-quality or worker-heterogeneity assumptions. Developing conditions that exploit without-replacement structure more sharply remains open.

\bibliography{references}
\bibliographystyle{ims}
\newpage

\appendix

\section{Proofs of the Main Results}
This section proves the main theoretical results. We first present two
algorithm-independent technical lemmas. \Cref{lem:alignment-deficit} is a
deterministic algebraic statement for arbitrary vectors.
\Cref{lem:one-step-signed-descent} applies pathwise to any sign-based update
under smoothness and a nonnegative stepsize. Neither lemma requires a
finite-sum objective, a particular sampling scheme, or an unbiased
estimator.
\begin{lemma}[Alignment deficit]\label{lem:alignment-deficit}
For every $\gb,\vb\in\RR^d$, the deficit in \eqref{eq:alignment-deficit} satisfies
\begin{align}
    0\leq \mathsf A(\gb,\vb)
    &\leq 2\sum_{j=1}^d |g_j|
    \ind\{\sign(v_j)\neq\sign(g_j)\},
    \label{eq:alignment-indicator-bound}\\
    \mathsf A(\gb,\vb)&\leq 2\lVert\vb-\gb\rVert_1.
    \label{eq:alignment-absolute-bound}
\end{align}
These inequalities include coordinates at which $g_j=0$ or $v_j=0$.
\end{lemma}
\begin{proof}
The $j$-th summand of $\mathsf A(\gb,\vb)$ is zero when $g_j=0$ or the two nonzero signs agree, is $|g_j|$ when $v_j=0\neq g_j$, and is $2|g_j|$ when the two signs are opposite. This proves nonnegativity and \eqref{eq:alignment-indicator-bound}. In the zero-estimator case, $|v_j-g_j|=|g_j|$; in the opposite-sign case, $|v_j-g_j|\geq |g_j|$. Hence each summand is at most $2|v_j-g_j|$, proving \eqref{eq:alignment-absolute-bound}.
\end{proof}

\begin{lemma}[Generic one-step signed descent]
\label{lem:one-step-signed-descent}
Let $f:\RR^d\to\RR$ have an $L$-Lipschitz gradient. For arbitrary
$\xb,\vb\in\RR^d$ and $\gamma\geq0$, define
$\xb^+:=\xb-\gamma\sign(\vb)$ and $\gb:=\nabla f(\xb)$. Then
\begin{align}
f(\xb^+)-f(\xb)
&\leq -\gamma\lVert\gb\rVert_1
  +\gamma\mathsf A(\gb,\vb)
  +\frac{dL\gamma^2}{2}
  \label{eq:one-step-alignment-descent}\\
&\leq -\gamma\lVert\gb\rVert_1
  +2\gamma\lVert\vb-\gb\rVert_1
  +\frac{dL\gamma^2}{2}.
  \label{eq:one-step-absolute-descent}
\end{align}
Moreover,
\begin{align}
  \gamma\mathsf A(\gb,\vb)
  \leq 2\gamma\sum_{j=1}^d |g_j|
  \ind\{\sign(v_j)\neq\sign(g_j)\}.
  \label{eq:one-step-indicator-descent}
\end{align}
\end{lemma}
\begin{proof}
By $L$-smoothness and $\lVert\sign(\vb)\rVert_2^2\leq d$,
\begin{align*}
    f(\xb^+)-f(\xb)
    &\leq -\gamma\langle\gb,\sign(\vb)\rangle
      +\frac{L\gamma^2}{2}\lVert\sign(\vb)\rVert_2^2\\
    &\leq -\gamma\lVert\gb\rVert_1
      +\gamma\mathsf A(\gb,\vb)
      +\frac{dL\gamma^2}{2}.
\end{align*}
The second inequality follows from \eqref{eq:alignment-absolute-bound} in
\Cref{lem:alignment-deficit}; the alternative bound follows from
\eqref{eq:alignment-indicator-bound} in
\Cref{lem:alignment-deficit}.
\end{proof}

\subsection{Proof of Theorem \ref{thm:signrr}}
\begin{proof}
For brevity, set
$\gb_t^i:=\nabla f(\xb_t^i)$ and
$\vb_t^i:=\nabla f_{\pi_i^t}(\xb_t^i)$. Applying
\eqref{eq:one-step-alignment-descent} in
\Cref{lem:one-step-signed-descent} pathwise with
$\xb=\xb_t^i$, $\vb=\vb_t^i$, and $\gamma=\gamma_t^i$, and then
rearranging, gives
\begin{align}
    \gamma_t^i\lVert\gb_t^i\rVert_1
    \leq f(\xb_t^i)-f(\xb_t^{i+1})
    +\frac{dL(\gamma_t^i)^2}{2}
    +\gamma_t^i\mathsf A(\gb_t^i,\vb_t^i).
    \label{eq:signrr-one-step-alignment}
\end{align}
This step uses neither independence nor division by a gradient coordinate, so it remains valid under random reshuffling and at zero coordinates. Taking expectations, summing over all $t$ and $i$, using $\xb_t^n=\xb_{t+1}^0$, and applying \Cref{ass:lower-bound} yield
\begin{align}
    \sum_{t=0}^{T-1}\sum_{i=0}^{n-1}\gamma_t^i
    \EE\lVert\gb_t^i\rVert_1
    \leq f(\xb_0)-f_*
    +\frac{dL}{2}\sum_{t=0}^{T-1}\sum_{i=0}^{n-1}(\gamma_t^i)^2
    +\sum_{t=0}^{T-1}\sum_{i=0}^{n-1}\gamma_t^i
    \EE\mathsf A(\gb_t^i,\vb_t^i).
    \label{eq:signrr-telescoped-alignment}
\end{align}
After dividing by $\sum_{t,i}\gamma_t^i$ and lower bounding the weighted average by its smallest term, we obtain
\begin{align}
    \min_{t,i}\EE\lVert\gb_t^i\rVert_1
    &\leq
    \frac{\sum_{t,i}\gamma_t^i\EE\lVert\gb_t^i\rVert_1}
         {\sum_{t,i}\gamma_t^i} \notag\\
    &\leq \frac{f(\xb_0)-f_*}{\sum_{t,i}\gamma_t^i}
    +\frac{dL}{2}\frac{\sum_{t,i}(\gamma_t^i)^2}{\sum_{t,i}\gamma_t^i}
    +\varepsilon_{\mathrm{align}}.
    \label{eq:signrr-general-weighted-bound}
\end{align}
For $\gamma_t^i=\gamma_0/\sqrt{nt+i+1}$, reindexing $k=nt+i+1$ gives the exact identities
\begin{align*}
    \sum_{t,i}\gamma_t^i=\gamma_0H_{N,1/2},
    \qquad
    \sum_{t,i}(\gamma_t^i)^2=\gamma_0^2H_N.
\end{align*}
Substitution into \eqref{eq:signrr-general-weighted-bound} proves \eqref{eq:signrr-alignment-bound}. Moreover,
\begin{align*}
    H_N\leq1+\log N,
    \qquad
    H_{N,1/2}\geq\int_1^{N+1}x^{-1/2}\,dx
    =2(\sqrt{N+1}-1),
\end{align*}
which proves \eqref{eq:signrr-explicit-bound}.

Finally, applying \eqref{eq:alignment-absolute-bound} in
\Cref{lem:alignment-deficit} pathwise and then taking expectations gives
$\varepsilon_{\mathrm{align}}\leq2\delta_{\mathrm{abs}}$. Conditional
Cauchy--Schwarz applied coordinatewise gives
\begin{align*}
    \EE\!\left[|[\vb_t^i-\gb_t^i]_j|\mid\xb_t^i\right]
    \leq
    \sqrt{\EE\!\left[[\vb_t^i-\gb_t^i]_j^2\mid\xb_t^i\right]},
\end{align*}
and averaging proves $\delta_{\mathrm{abs}}\leq\sigma_{\mathrm{rms}}$ and hence \eqref{eq:alignment-error-hierarchy}. Because the derivation of
\eqref{eq:signrr-telescoped-alignment} uses only
\eqref{eq:one-step-alignment-descent} in
\Cref{lem:one-step-signed-descent}, not the diminishing schedule, setting all
stepsizes equal to $\gamma$ and dividing by $nT\gamma$ proves the
constant-stepsize claim in \Cref{cor:signrr-constant}.
\end{proof}

\subsection{Proof of Proposition \ref{prop:signrr-impossibility}}
\begin{proof}
For every $x\in(-2,1)$,
\begin{align*}
    \sign(f_0'(x))=1,
    \qquad
    \sign(f_1'(x))=-1,
    \qquad
    f'(x)=x+\frac12.
\end{align*}
Set $x_t:=x_t^0$ and write $\eta_k:=\gamma_0/\sqrt{k+1}$, so that
$\gamma_t^i=\eta_{2t+i}$. Define
\begin{align*}
\xi_t:=
\begin{cases}
 +1,&(\pi_0^t,\pi_1^t)=(0,1),\\
 -1,&(\pi_0^t,\pi_1^t)=(1,0).
\end{cases}
\end{align*}
By the independent uniform epoch permutations in \Cref{alg:signrr}, the
sequence $\{\xi_t\}_{t\geq0}$ consists of independent Rademacher variables. Let
$\delta_t:=\eta_{2t}-\eta_{2t+1}>0$. Whenever $x_t\in(-2,1)$, the first
selected component has gradient sign $\xi_t$, so the first update gives
\begin{align}
    x_t^1
    &=x_t-\eta_{2t}\xi_t,
    \label{eq:counterexample-inner-recurrence}
\end{align}
If also $x_t^1\in(-2,1)$, the second selected component has gradient sign
$-\xi_t$, and the second update gives
\begin{align}
    x_{t+1}=x_t^2
    &=x_t^1+\eta_{2t+1}\xi_t
      =x_t-\delta_t\xi_t.
    \label{eq:counterexample-epoch-recurrence}
\end{align}
Moreover,
\begin{align*}
    \sum_{t=0}^{\infty}\delta_t
    &=\gamma_0\sum_{t=0}^{\infty}
      \left(\frac1{\sqrt{2t+1}}-\frac1{\sqrt{2t+2}}\right)\\
    &\leq\gamma_0\sum_{k=1}^{\infty}
      \left(\frac1{\sqrt{k}}-\frac1{\sqrt{k+1}}\right)
    =\gamma_0.
\end{align*}
For $D_t:=\sum_{s=0}^{t-1}\delta_s$, with $D_0:=0$, we now prove by induction that
$|x_t|\leq D_t\leq\gamma_0$. This is true at $t=0$. If it holds at epoch
$t$, then $x_t\in(-2,1)$, so the first selected component has sign $\xi_t$
and \eqref{eq:counterexample-inner-recurrence} holds. Consequently,
\begin{align*}
    |x_t^1|\leq D_t+\eta_{2t}\leq2\gamma_0\leq\frac12,
\end{align*}
so $x_t^1\in(-2,1)$ as well. The second selected component therefore has
sign $-\xi_t$, which proves \eqref{eq:counterexample-epoch-recurrence} and
$|x_{t+1}|\leq D_t+\delta_t=D_{t+1}$. This completes the induction and
also controls the inner checkpoint $x_t^1$, where the second component
gradient is evaluated.

With $\mathcal H_t=\sigma(\pi^0,\ldots,\pi^{t-1})$ as defined above,
$x_t$ is $\mathcal H_t$-measurable, whereas $\xi_t$ is independent of
$\mathcal H_t$ and has mean zero. The two recurrences above imply
\begin{align*}
    \EE[x_{t+1}\mid\mathcal H_t]&=x_t,\\
    \EE[x_t^1\mid\mathcal H_t]&=x_t.
\end{align*}
Since $x_0=0$, induction and the tower property give
$\EE[x_t]=\EE[x_t^0]=\EE[x_t^1]=0$ for every $t$. For each
$i\in\{0,1\}$, the bounds above imply $x_t^i\geq-1/2$, so
\begin{align*}
    \EE|f'(x_t^i)|
    =\EE\left[x_t^i+\frac12\right]
    =\frac12,
\end{align*}
which proves \eqref{eq:signrr-counterexample-gradient}.
\end{proof}

\subsection{Proof of Proposition \ref{prop:aggregate-alignment}}
\begin{proof}
The pathwise derivation of \eqref{eq:signrr-telescoped-alignment} from
\eqref{eq:one-step-alignment-descent} in
\Cref{lem:one-step-signed-descent} is valid for arbitrary positive
deterministic stepsizes. Substitute \eqref{eq:aggregate-relative-alignment}
into that inequality and move the resulting
$(1-\rho)$ multiple of the weighted gradient sum to the left. This gives
\begin{align*}
    \rho\sum_{t,i}\gamma_t^i
      \EE\lVert\nabla f(\xb_t^i)\rVert_1
    \leq f(\xb_0)-f_*+\frac{dL}{2}\sum_{t,i}(\gamma_t^i)^2+B_N.
\end{align*}
Divide by $\rho\sum_{t,i}\gamma_t^i$ and lower bound the weighted average
by its minimum.
\end{proof}

\subsection{Proof of Theorem \ref{thm:signrr-aligned}}
\begin{proof}
Conditioning the smoothness inequality on $\mathcal F_t^i$ and using
\eqref{eq:remaining-sign-field} gives
\begin{align*}
    \EE\!\left[f(\xb_t^{i+1})\mid\mathcal F_t^i\right]
    &\leq f(\xb_t^i)
      -\gamma_t^i
       \left\langle\nabla f(\xb_t^i),\mathbf{s}_t^i\right\rangle
      +\frac{dL(\gamma_t^i)^2}{2}\\
    &\leq f(\xb_t^i)
      -\rho\gamma_t^i\lVert\nabla f(\xb_t^i)\rVert_1
      +\frac{dL(\gamma_t^i)^2}{2},
\end{align*}
where the last line uses \Cref{ass:remaining-alignment}. Taking total
expectations, telescoping over all inner steps and epochs, and using
$f(\xb_T)\geq f_*$ yield
\begin{align*}
    \rho\sum_{t,i}\gamma_t^i
      \EE\lVert\nabla f(\xb_t^i)\rVert_1
    \leq f(\xb_0)-f_*+\frac{dL}{2}\sum_{t,i}(\gamma_t^i)^2.
\end{align*}
Division by $\rho\sum_{t,i}\gamma_t^i$ proves
\eqref{eq:signrr-aligned-general}.  The harmonic-sum identities used in the
proof of \Cref{thm:signrr} give \eqref{eq:signrr-aligned-diminishing}.
For a constant step, the bound is
\begin{align*}
    \frac{f(\xb_0)-f_*}{\rho N\gamma}+\frac{dL\gamma}{2\rho}.
\end{align*}
Minimizing this expression over $\gamma>0$ gives the stepsize in
\Cref{cor:signrr-constant} and proves \eqref{eq:signrr-aligned-optimized}.
\end{proof}

\subsection{Proof of Theorem \ref{thm:signrvr}}
\begin{proof}
Fix an epoch $t$ and abbreviate
$\gb_t^i:=\nabla f(\xb_t^i)$ and $\vb_t^i$ as in
\eqref{eq:signrvr-estimator}. By the triangle inequality and
\Cref{ass:smoothness},
\begin{align}
\lVert\vb_t^i-\gb_t^i\rVert_2
&\leq
\lVert\nabla f_{\pi_i^t}(\xb_t^i)-\nabla f_{\pi_i^t}(\yb_t)\rVert_2
+\lVert\nabla f(\yb_t)-\nabla f(\xb_t^i)\rVert_2 \notag\\
&\leq 2L\lVert\xb_t^i-\yb_t\rVert_2.
\label{eq:signrvr-estimator-error}
\end{align}
The absolute-error bound \eqref{eq:alignment-absolute-bound} in
\Cref{lem:alignment-deficit} and
$\lVert\zb\rVert_1\leq\sqrt d\lVert\zb\rVert_2$ therefore give
\begin{align}
\mathsf A(\gb_t^i,\vb_t^i)
&\leq2\lVert\vb_t^i-\gb_t^i\rVert_1
\leq4\sqrt dL\lVert\xb_t^i-\yb_t\rVert_2.
\label{eq:signrvr-alignment-distance}
\end{align}
Because $\yb_t=\xb_t^0$ and
$\lVert\sign(\vb_t^r)\rVert_2\leq\sqrt d$, the inner-loop updates imply
\begin{align}
\lVert\xb_t^i-\yb_t\rVert_2
&\leq\sum_{r=0}^{i-1}\gamma_t^r
\lVert\sign(\vb_t^r)\rVert_2
\leq\sqrt d\sum_{r=0}^{i-1}\gamma_t^r.
\label{eq:signrvr-epoch-drift}
\end{align}
Combining \eqref{eq:signrvr-alignment-distance} and
\eqref{eq:signrvr-epoch-drift} yields
\begin{align}
\mathsf A(\gb_t^i,\vb_t^i)
\leq4dL\sum_{r=0}^{i-1}\gamma_t^r.
\label{eq:signrvr-alignment-step}
\end{align}

Applying \eqref{eq:one-step-alignment-descent} in
\Cref{lem:one-step-signed-descent} with $\xb=\xb_t^i$,
$\vb=\vb_t^i$, and $\gamma=\gamma_t^i$, substituting
\eqref{eq:signrvr-alignment-step}, and rearranging gives
\begin{align}
\gamma_t^i\lVert\gb_t^i\rVert_1
&\leq f(\xb_t^i)-f(\xb_t^{i+1})
+\frac{dL}{2}(\gamma_t^i)^2
+4dL\gamma_t^i\sum_{r=0}^{i-1}\gamma_t^r.
\label{eq:signrvr-one-step}
\end{align}
For arbitrary real numbers $a_i$,
\begin{align}
\sum_{i=0}^{n-1}a_i\sum_{r=0}^{i-1}a_r
=\frac12\left[\left(\sum_{i=0}^{n-1}a_i\right)^2
-\sum_{i=0}^{n-1}a_i^2\right].
\label{eq:cross-sum-identity}
\end{align}
Using this identity with $a_i=\gamma_t^i$ and summing
\eqref{eq:signrvr-one-step} over an epoch gives
\begin{align}
\sum_{i=0}^{n-1}\gamma_t^i\lVert\gb_t^i\rVert_1
&\leq f(\xb_t)-f(\xb_{t+1})
+2dLS_t^2-\frac{3dL}{2}Q_t.
\label{eq:signrvr-one-epoch}
\end{align}
Summing over $t$ and using $f(\xb_T)\geq f_*$ proves the corresponding
pathwise inequality without the expectation on its left-hand side. Taking
expectations proves the first inequality in \eqref{eq:signrvr-general}, and
the second follows from $Q_t\geq0$. Dividing by $\sum_t S_t$ and lower
bounding the weighted average by its smallest term proves
\eqref{eq:signrvr-min-general}.

It remains to specialize the stepsizes. For
$\gamma_t^i=\gamma_0/\sqrt{nt+i+1}$,
\begin{align*}
S_0&=\gamma_0H_{n,1/2}\leq2\gamma_0\sqrt n,\\
S_t&\leq\frac{n\gamma_0}{\sqrt{nt}}
=\gamma_0\sqrt{\frac nt},\qquad t\geq1.
\end{align*}
Hence
\begin{align}
\sum_{t=0}^{T-1}S_t^2
\leq n\gamma_0^2\bigl(4+H_{T-1}\bigr),
\label{eq:signrvr-diminishing-epoch-sum}
\end{align}
while $\sum_t S_t=\gamma_0 H_{N,1/2}$. Dropping the negative $Q_t$ term in
\eqref{eq:signrvr-min-general} proves \eqref{eq:signrvr-diminishing}.

Finally, if every stepsize equals $\gamma$, then
$S_t=n\gamma$ and $Q_t=n\gamma^2$. Dividing
\eqref{eq:signrvr-general} by $nT\gamma$ gives
\eqref{eq:signrvr-constant}. The two terms on its right-hand side are
minimized at the stated positive stepsize when $\Delta_0>0$; substitution
gives \eqref{eq:signrvr-optimized}.
\end{proof}

\end{document}